\documentclass{article}

\usepackage{microtype}
\usepackage{graphicx}
\usepackage{subfigure}
\usepackage{booktabs} 

\usepackage{hyperref}

\usepackage[accepted]{icml2024}

\usepackage{amsmath}
\usepackage{amssymb}
\usepackage{mathtools}
\usepackage{amsthm}
\usepackage{lipsum}

\usepackage[capitalize,noabbrev]{cleveref}

\theoremstyle{plain}

\theoremstyle{definition}

\theoremstyle{remark}

\usepackage[textsize=tiny]{todonotes}

\newcommand{\txn}{\mathcal{T}}

\renewcommand{\>}{\rangle}

\DeclareMathOperator{\safesach}{SafeSAC-H}

\icmltitlerunning{Safe Reinforcement Learning with Learned Non-Markovian Safety Constraints}

\begin{document}

\twocolumn[
\icmltitle{Safe Reinforcement Learning with Learned Non-Markovian Safety Constraints}




\begin{icmlauthorlist}
\icmlauthor{Siow Meng Low}{smu}
\icmlauthor{Akshat Kumar}{smu}
\end{icmlauthorlist}

\icmlaffiliation{smu}{School of Computing and Information Systems, Singapore Management University, Singapore}

\icmlcorrespondingauthor{Siow Meng Low}{smlow.2020@phdcs.smu.edu.sg}

\icmlkeywords{Machine Learning, Deep Reinforcement Learning, Safe AI, ICML}

\vskip 0.3in
]




\begin{abstract}
In safe Reinforcement Learning (RL), safety cost is typically defined as a function dependent on the immediate state and actions. In practice, safety constraints can often be non-Markovian due to the insufficient fidelity of state representation, and safety cost may not be known. We therefore address a general setting where safety labels (e.g., safe or unsafe) are associated with state-action \textit{trajectories}. Our key contributions are: \textit{first}, we design a safety model that specifically performs credit assignment to assess contributions of partial state-action trajectories on safety. This safety model is trained using a labeled safety dataset. \textit{Second}, using RL-as-inference strategy we derive an effective algorithm for optimizing a safe policy using the learned safety model. \textit{Finally}, we devise a method to dynamically adapt the tradeoff coefficient between reward maximization and safety compliance. We rewrite the constrained optimization problem into its dual problem and derive a gradient-based method to dynamically adjust the tradeoff coefficient during training. Our empirical results demonstrate that this approach is highly scalable and able to satisfy sophisticated non-Markovian safety constraints.
\end{abstract}

\section{Introduction}
\label{intro}

With the increasing deployment of autonomous agent in critical real-world application areas such as autonomous vehicle~\cite{kiran2021deep}, smart heating control~\cite{gupta2021energy} and energy management~\cite{zhang2019deep}, safety has become an important research area in RL~\cite{garcia2015comprehensive} to ensure that the deployed autonomous agents behave safely. One popular safe RL method adopts constrained MDP framework~\cite{altman1998constrained} to limit the cumulative safety cost incurred along the trajectories traversed by autonomous agents~\cite{tessler2018reward, stooke2020responsive, ha2021learning}. Other safe RL approaches include learning to avoid actions which potentially land the agent into a set of unsafe states~\cite{thomas2021safe, thananjeyan2021recovery}, constraining the valid actions to be performed by agent~\cite{lin2021escaping, kasaura2023benchmarking}, and specifying safety constraints using logic specifications~\cite{alshiekh2018safe, jansen2020safe, jothimurugan2021compositional}, such as temporal logic. 

A major difficulty in applying the constrained RL approach~\cite{achiam2017constrained, tessler2018reward, stooke2020responsive} is the difficulty in correctly specifying the safety cost function, especially in terms of numerical values. It is well recognized that reward specification design is challenging~\cite{ng2000algorithms, ray2019benchmarking} and incorrect specification can lead to unintended agent behaviors. There is no reason why safety cost design is any less challenging, especially when there is uncertainty around what constitutes a safe behavioral pattern. Similar to constrained RL, all the other safe RL methods mentioned assume in-depth knowledge on the safety requirements. 

To address this problem of having incomplete knowledge on safety, a number of research study has been conducted to learn the safety cost function from human feedback~\cite{SaisubramanianZ21, zhang2018minimax}, infer implicit human preferences from initial state configuration~\cite{shah2019preferences} and nudge agent in preserving flexibility metrics such as reachability measure or attainable utility~\cite{krakovna2018penalizing, turner2020avoiding, turner2020conservative}. Another line of related work is ICRL - inverse constrained reinforcement learning~\cite{malik2021inverse, gaurav2022learning, liu2022benchmarking} where constraints are inferred from expert demonstration data. The learned constraints are expressed in the form of feasibility variables to quantify the feasibility of a state-action pair. 

One obstacle in applying the ICRL methods is the collection of expert demonstration. It can be challenging to demonstrate what an expert would do, especially when the task involves transition dynamics which is complex and unknown. In contrast, it is relatively easier for a human safety designer or computer program to recognize unsafe patterns in a trajectory. Another shortcoming of ICRL is that it encodes the constraint as feasibility score which depends only on state-action pair. Due to the MDP designer's incomplete knowledge on safety, it is highly likely that the state specification lacks sufficient fidelity~\cite{saisubramanian2022avoiding} for the safety constraint to be modeled as Markovian. Considering this, we argue that the safety model should accommodate non-Markovian constraint.

In this work, we study how non-Markovian constraint can be inferred from labeled data, and subsequently be used to facilitate safe RL. Our main contributions are summarized as follows. \textit{Firstly}, we design a unique safety model to learn non-Markovian constraint from labeled data. Unlike other methods, our safety model preserves the non-Markovian relationship by learning the embedding to encode trajectory history. \textit{Secondly}, we perform probabilistic inference and derive a model-free RL method which optimizes total reward while respecting the learned non-Markovian constraint. This method can be seen as an extension to off-policy Soft Actor-Critic (SAC) method~\citep{haarnoja2018soft} and we name it $\safesach$. \textit{Thirdly}, we devise a method to automatically adjust the Lagrange multiplier to ensure constraint compliance. This method removes the hurdle for practitioners in applying this method since it can be tricky and tedious to manually adjust the Lagrange multiplier. \textit{Lastly}, our empirical results on a variety of continuous control domains demonstrate the applicability of our method to a wide range of safe RL problems with unknown non-Markovian safety constraint.

\section{Problem Formulation}

\subsection{Constrained Reinforcement Learning} \label{section:crl}

Markov Decision Processes (MDPs)~\cite{sutton1998introduction} is widely used to model sequential decision making problem under uncertainty. Constrained reinforcement learning (CRL) methods utilize an extended MDP model, called Constrained Markov Decision Processes (CMDPs)~\cite{altman1998constrained}. CMDPs allow specifying safety requirements as constraints in the problem formulation; CRL algorithms seek policy which optimizes reward while respecting the specified safety constraints.  

A CMDP is defined by tuple $(S, A, \txn, R, b_0, C, d, T)$. We consider a general setting with continuous state and action spaces ($S\subseteq \mathbb{R}^n$, $A\subseteq \mathbb{R}^m$). The environment state transition is characterized by the function $p(s_{t+1}| s_t, a_t) \!=\!\txn(s_t, a_t, s_{t+1})$. The reward function $R: S \times A \rightarrow \mathbb{R}$ maps a state, action to a scalar reward value. Similarly, a constraint function $C: S \times A \rightarrow \mathbb{R}$ maps a state, action to a scalar cost value and $d$ is the associated bound on this cost, and $b_0$ is the initial state distribution. Lastly, $T$ refers to the planning horizon. 

A CRL algorithm solves for a policy $\pi$ (parameterized by $\theta$) which maximizes the expected sum of rewards subject to the specified constraint:
\begin{equation}
\begin{aligned} \label{eq:crl}
	&\max_{\theta}  J_{R}(\pi_{\theta}) = \mathbb{E}_{\tau \sim \pi_{\theta}} [ \sum_{t=0}^{T} 
 R(s_t, a_t) ] \\
	&\text{s.t.} \quad J_{c}(\pi_{\theta}) = \mathbb{E}_{\tau \sim \pi_{\theta}} [\sum_{t=0}^{T} 
 C(s_t, a_t)] \leq d
\end{aligned}
\end{equation}
The above optimization program exposes two limitations of the existing CMDP formulation: (1) The cost function $C$ is Markovian and only depends on current state-action pair $(s, a)$; (2) numerical values for $C(s, a)$ are given by the environment simulator. In a more general setting, neither of these assumptions may hold. It can be difficult for safety designer to quantify the numerical safety costs. Furthermore, modelling the safety cost as the additive sum of Markovian costs may not be feasible if the state specification in CMDP is incomplete~\cite{saisubramanian2022avoiding}. In this case, safety has to be modeled as a function of the entire trajectory $\tau$, instead of only dependent on immediate state-action pair. To address such limitations, we propose a safety model which predicts whether a given trajectory is safe, and learn it using a dataset consisting of variable-length trajectory segments and their respective safety labels (i.e. safe / unsafe). This avoids the need to specify numerical costs for safety, and safety is dependent on a state-action trajectory.



\begin{figure}[t]
    \centering
    \includegraphics[width=0.49\textwidth]{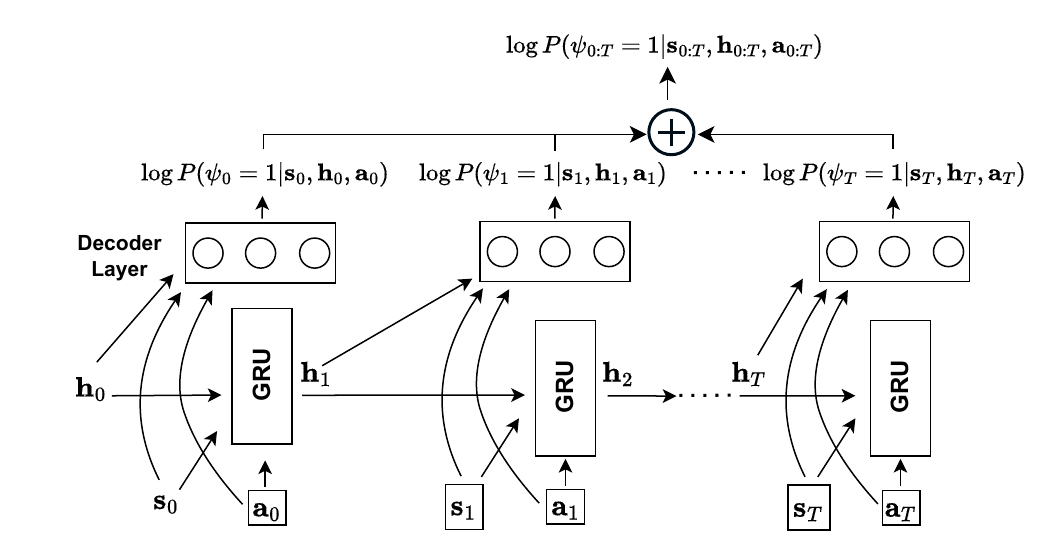}
    \vskip 0pt
    \caption{Non-Markovian Safety Model}
    \label{fig:safety_model}
    \vskip 0pt
\end{figure}

\subsection{Trajectory Safety} \label{section:traj_safety}

We define safety as observed over a (sub)trajectory of variable length $t\!+\!1$. We denote this trajectory as $\tau_{0:t}\!=\!\<s_0, a_0, s_1, \ldots, s_{t}, a_{t}\>$, comprising of $t\!+\!1$ state-action pairs. A binary label $\psi$ indicates whether the observed trajectory $\tau_{0:t}$ is safe or unsafe. It can be viewed as an indicator function where $\Psi(\tau_{t_1:t_2}) = 1$ indicates $\tau_{t_1:t_2}$ is safe; $\Psi(\tau_{t_3:t_4}) = 0$ points to an unsafe trajectory segment $\tau_{t_3:t_4}$.

For a trajectory $\tau_{0:t}$ to be safe ($\Psi(\tau_{0:t}) \!=\! 1$), all the sub-trajectory segments contained within it must be safe, i.e. $\Psi(\tau_{t_1:t_2}) = 1, \forall (t_1, t_2), 0 \leq t_1 \leq t_2 \leq t$. Conversely, a trajectory $\Psi(\tau_{0:t}) = 0$ is unsafe if there is at least one unsafe sub-trajectory segment within it, i.e. $\exists\, \Psi(\tau_{t_1:t_2}) = 0, 0 \leq t_1 \leq t_2 \leq t$. 

Note that the safety labeling function is unknown in general and it has to be learned from data. A dataset of labeled (sub)trajectories $\{(\tau^{i}_{t_1:t_2}, \psi)\}$ is used to facilitate learning of such a safety model. 
To infer the non-Markovian safety constraints and subsequently enable safe RL, such safety dataset has to be collected beforehand. For our experiments, we store the observed trajectory data seen during the learning process of unconstrained RL. This approach is aligned with incremental learning process, where the historical trajectory data is collected and used for safe RL training subsequently. In our experiments, we assign safety labels using a non-Markovian safety criterion, but this criterion is not provided to the safety model or the safe RL algorithm.

\section{Learning Non-Markovian Safety Model}

To learn the safety labeling function for a variable-length trajectory segment, we designed a recurrent neural network~\cite{rumelhart1986learning}. The network architecture of the RNN model is depicted in Figure~\ref{fig:safety_model} and gated recurrent units (GRU) is used as it has been shown to achieve a good balance between accuracy and network parameter size~\cite{cho2014properties}. 

The inputs to this safety model consist of: (1) state-action trajectory $\tau_{0:T}\!=\!\langle s_0, a_0, s_1, a_1 \ldots, s_{T}, a_{T}\rangle$ observed in the trajectory segment; (2) all-zero vector $h_0$ to signify the beginning of a trajectory segment. At a timestep $t$, the GRU unit summarizes information about current state-action pair $(s_t, a_t)$ and hidden vector $h_t$ into the next hidden vector $h_{t+1}$. The same tuple $(s_t, h_t, a_t)$ is also passed into a decoder layer to produce a log-probability scalar value $\log P(\psi_t = 1 | s_t, h_t, a_t)$, which can be interpreted as log probability of being safe after observing the current state-action pair $(s_t, a_t)$ and hidden vector $h_t$. We next define the concrete connection between a trajectory $\tau_{0:T}$ being safe and different $\log P(\psi_t | \cdot)$ values.

This architecture implies that variables $\psi_t$ and $\psi_{t'}$ are conditionally independent given $(s_t, h_t, a_t)$ and $(s_{t'}, h_{t'}, a_{t'})$.
Consequently, the log-probability of all timesteps being jointly safe is the sum of all log-probabilities, i.e. 
\vskip -1pt
{\small
\begin{align}
&\log P(\Psi(\tau_{0:T})\!=\!1\!) \!=\!\log P(\psi_{0} \!=\! 1,\! \cdots\!, \psi_{T} \!=\! 1 |\! s_{0:T}, h_{0:T}, a_{0:T}\!) \nonumber \\
&= \sum_{t=0}^{T} \log P(\psi_t = 1 | s_t, h_t, a_t)
\end{align}}
We shall later use the shorthand: 
\vskip -1pt
{\small
\begin{align}
P(\psi_{0:T} = 1 | s_{0:T}, h_{0:T}, a_{0:T}) = \prod_{i=0}^T P(\psi_{i} = 1 | s_{i}, h_{i}, a_{i})
\end{align}}
Since the output scalar of this safety model is a log-probability, binary cross entropy loss can be used to train the safety model to differentiate between safe and unsafe trajectories. 

The proposed safety model adheres to the safety definition introduced in Section~\ref{section:traj_safety}. Recall that for a trajectory to be safe, every single trajectory segment contained within it must be safe. For such a trajectory, the safety model is trained to output log-probability scores close to zero for all timesteps. This ensures that the final joint log-probability $\log P(\psi_{0:T} = 1 | s_{0:T}, h_{0:T}, a_{0:T})$ sums to a near-zero value, which indicates that the trajectory is predicted to be safe with high confidence. For the case of unsafe trajectory, highly negative log-probability $\log P(\psi_t = 1 | s_t, h_t, a_t)$ at even a single timestep $t$ can pull down the final sum to a highly negative number. This causes the safety model to predict the trajectory to be unsafe with high confidence. This is aligned with the unsafe trajectory definition: if there exists at least a single unsafe sub-segment, the whole trajectory is deemed as unsafe. 

The novelty of this architecture is that it decomposes the predicted log-probability score (of a trajectory being safe) into a sum of multiple log-probability scores, one for each timestep. This enables us to apply the conventional RL techniques in solving non-Markovian CRL (as shown in Section~\ref{section:nmcrl}). 

One final comment is that the trajectory classification method presented above allows a modular separation between optimizing the primary task of the underlying MDP, and safety specifications. As more safety data is gathered, we can update our safety model with additional data, without the need to always change the underlying state space 
each time new unsafe patterns are discovered. This enables modular and continual safety learning, which is critical for real world deployment of autonomous agents.

\section{Non-Markovian Constrained RL} \label{section:nmcrl}

\subsection{Non-Markovian CRL with Safety Model} \label{section:safety_model}

Once trained, the safety model extracts information on non-Markovian safety constraints from the labeled trajectory data. It outputs low log-probability score for trajectories exhibiting non-Markovian unsafe patterns. Conversely, high log-probability score will be predicted for safe trajectories. With $\phi$ being the parameters of the safety model, the set of output log-probabilities $\log P_{\phi}(\psi_t = 1 | s_t, h_t, a_t)$ allows us to rewrite the non-Markovian CRL program in a similar manner as the conventional CRL program:
\vskip 0pt
{\small
\begin{equation} \label{eq:nmcrl}
\begin{aligned}
 \max_{\theta} & \: J_{R}(\pi_{\theta}) = \mathbb{E}_{\tau \sim \pi_{\theta}} [ \sum_{t=0}^{T} R(s_t, a_t) ] \\
  \textrm{s.t.} & \: \mathbb{E}_{\tau \sim \pi_{\theta}} [ \exp(\sum_{t=0}^{T} \log P_{\phi}({\psi}_t = 1 | s_t, h_t, a_t)) ] \geq d
\end{aligned}
\end{equation}}
The goal of this non-Markovian CRL program is to maximize the sum of rewards, subject to the condition that the proportion of safe trajectories is above threshold $d$, with a valid range of $[0, 1]$. We also note that the hidden vector $h_t$ is a deterministic output of the safety model, provided the input $(s_{t-1}, h_{t-1}, a_{t-1})$ is given. Therefore, all the hidden vectors can be computed using the safety model for a trajectory $\tau$. 

We also consider the policy to depend on both $s$ and $h$, denoted as $\pi_{\theta}(a | s, h)$. By observing $h$, the policy gains information about the trajectory history and can take this into account while deciding the next action. This empowers the policy to better comply with the non-Markovian constraint. In our empirical study (Section~\ref{section:h_ablation}), we consider a conventional policy $\pi_{\theta}(a | s)$, which does not depend on history summary vector $h$, and show that it is not able to fully comply with non-Markovian constraint.

\subsection{Non-Markovian CRL as Probabilistic Inference}

\subsubsection{Probabilistic Graphical Model}

To solve the CRL problem in~\eqref{eq:nmcrl}, we derive an iterative policy improvement algorithm using probabilistic inference. The first step is to reformulate the CRL problem as a probabilistic inference problem by modeling the relationship between states, actions and hidden states in a probabilistic graphical model (PGM). Using the optimality variable concept introduced in~\citet{levine2018reinforcement}, we use two sets of optimality variables $\mathcal{O}_t$ and $\psi_t$ as shown in Figure~\ref{fig:graphical_model}. Both $\mathcal{O}_t$ and $\psi_t$ are binary random variables, where $\mathcal{O}_t = 1$ denotes that optimum reward is achieved at timestep $t$, and $\psi_t = 1$ indicates optimum safety score is achieved at timestep $t$. We define the distribution of these two variables as:
\begin{figure}[t]
    \centering
    \includegraphics[width=0.49\textwidth]{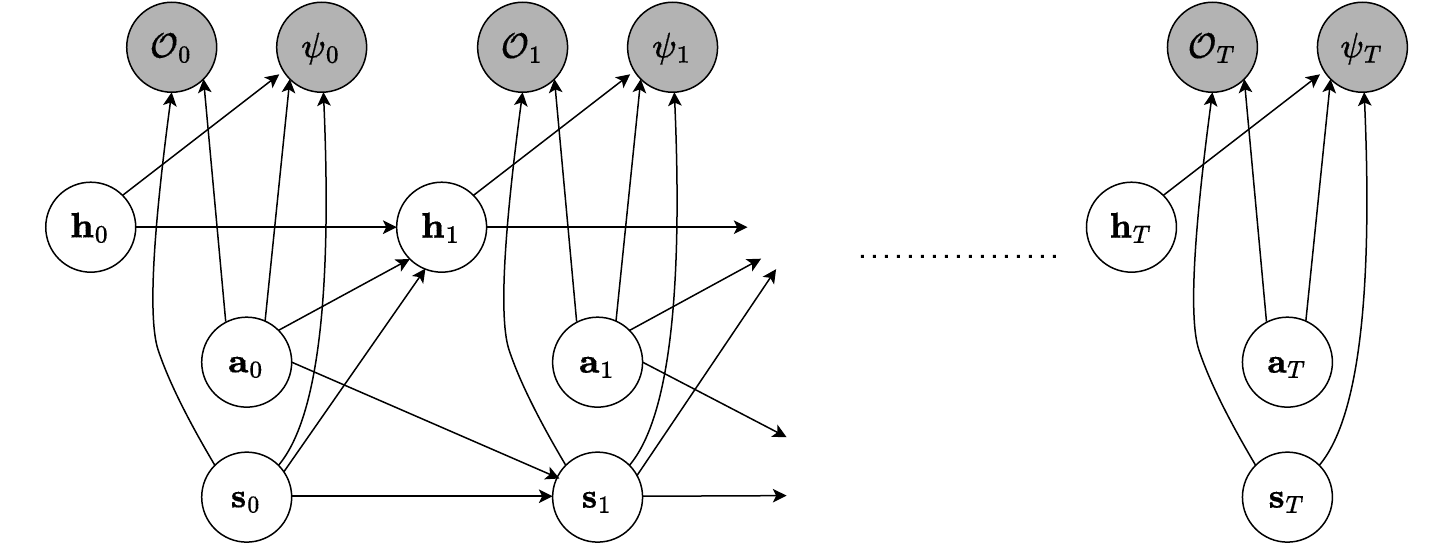}
    \vskip 0pt
    \caption{Graphical Model with two sets of optimality variables}
    \label{fig:graphical_model}
    \vskip 0pt
\end{figure}
\begin{equation} 
\small 
\label{eq:optimality_variables}
\begin{aligned}
    P(\mathcal{O}_t = 1 | s_t, a_t) = \exp(r(s_t, a_t)) \\
    P({\psi}_t = 1 | s_t, h_t, a_t) = {P_{\phi}({\psi}_t} = 1 | s_t, h_t, a_t) 
\end{aligned}
\end{equation}
With ${P_{\phi}({\psi}_t} = 1 | s_t, h_t, a_t)$ being the probability provided by the safety model trained in Section~\ref{section:safety_model}. 

To perform variational inference, we first obtain the probability of observing a trajectory $\tau$, conditioning on the optimality conditions---$\mathcal{O}_t = 1 \text{ and } {\psi}_t = 1 \; \forall t \in [0\!:\!T]$. We use the shorthand $\boldsymbol{o}_{0:T}$ for $\mathcal{O}_t = 1 \forall t \in [0:T]$ and define analogously $\boldsymbol{\psi}_{0:T}$.

This optimal distribution $p(\tau | \boldsymbol{o}_{0:T}, \boldsymbol{\psi}_{0:T})$ is derived as: 
{\small
\begin{equation} \label{eq:optimal_tau}
\begin{aligned}
    & p(\tau | \boldsymbol{o}_{0:T}, \boldsymbol{\psi}_{0:T}) \propto p(\tau, \boldsymbol{o}_{0:T}, \boldsymbol{\psi}_{0:T}) \\
    = & p({s}_0) \prod_{t=0}^{T-1} [ p({s}_{t+1} | {s}_{t}, {a}_{t}) \exp(r({s}_t, {a}_t)) p({\psi}_t = 1 | {s}_{t}, {h}_{t}, {a}_{t}) ] \\
    & \exp(r({s}_T, {a}_T)) p({\psi}_T = 1 | {s}_{T}, {h}_{T}, {a}_{T})
\end{aligned}
\end{equation}}
We refer the readers to Appendix~\ref{appendix:proof_p_tau} for full derivation. 

\subsubsection{Variational Inference} \label{section:var_inf}

Similar to how we derive the optimal trajectory distribution in Equation~\eqref{eq:optimal_tau}, we can also express the distribution of observed trajectory $\hat{p}(\tau; \pi)$ by following policy $\pi$. This paper considers a probabilistic policy $\pi(a | s, h)$, which represents the probability of executing action $a$ after having observed state $s$ and hidden state $h$. Essentially, safety model summarizes the history into $h$ and the policy decides an action after observing the state $s$ and the summary vector $h$.

The distribution of observed trajectory $\hat{p}(\tau; \pi)$ by following this policy $\pi(a | s, h)$ is expressed as: 
\begin{equation} \label{eq:approx_tau}
\small
\begin{aligned}
    \hat{p}(\tau; \pi) \!=\! p({s}_0) \prod_{t=0}^{T-1} [ p({s}_{t+1} | {s}_{t}, {a}_{t}) \pi({a_t} | {s_t}, {h_t}) ] \pi({a_T} | {s_T}, {h_T})
\end{aligned}
\end{equation}
As shown in~\cite{levine2018reinforcement}, a good control strategy can be found by maximizing the negative KL divergence between the distribution $\hat{p}(\tau; \pi)$ and optimal distribution $p(\tau | \boldsymbol{o}_{0:T}, \boldsymbol{\psi}_{0:T})$. Thus, the variational inference problem is:
\begin{equation}
\small
\begin{aligned}
    \max_{\hat{p}} -D_{KL}(\hat{p}(\tau; \pi) || p(\tau, \boldsymbol{o}_{0:T}, \boldsymbol{\psi}_{0:T}))
\end{aligned}
\label{eq:varObj}
\end{equation}
with the negative KL divergence $-D_{KL}$ as:
\begin{equation} \label{eq:kl_divergence}
\small
\begin{aligned}
    & -D_{KL}(\hat{p}(\tau; \pi) || p(\tau, \boldsymbol{o}_{0:T}, \boldsymbol{\psi}_{0:T})) \\
    = &\mathbb{E}_{\hat{p}(s_{0:T}, a_{0:T})} \Big [ \sum_{t=0}^{T} \big[r({s}_t, {a}_t) + \log p({\psi}_t = 1 | {s}_{t}, {h}_{t}, {a}_{t}) \\
    & - \log \pi({a_t} | {s_t}, {h_t})\big] \Big ]
\end{aligned}
\end{equation}
We refer the readers to Appendix~\ref{appendix:kl_divergence} for full derivation in arriving at the final KL-Divergence expression. 

Now consider a time step $t' \leq T$, to solve for policy $\pi(a_{t'} | s_{t'}, h_{t'})$, we only need to optimize the terms containing timestep $t'$ to $T$. This is because an action at timestep $t'$ does not affect the terms preceeding $t'$. The negative KL divergence containing only the terms from timestep $t'$ to $T$ is: 
\begin{equation} \label{eq:var_lb}
\small
\begin{aligned}
    \mathbb{E}_{\hat{p}(s_{t':T}, a_{t':T}; \pi)} \Big [ {}& \sum_{t=t'}^{T} [r({s}_t, {a}_t) - \log \pi({a_t} | {s_t}, {h_t}) \\
    & + \log p({\psi}_t = 1 | {s}_{t}, {h}_{t}, {a}_{t})] \Big ]
\end{aligned}
\end{equation}
To optimize the above expression, we employ a dynamic programming approach. To derive the recursive dynamic programming-style expressions, we define two Q-functions $Q_R$ and $Q_{\psi}$ which are both recursive functions:
\begin{equation} \label{eq:Q_R_function}
\small
\begin{aligned}
    {}& Q_R(s_t, h_t, a_t) \\
    = & r(s_{t}, a_{t}) + \mathbb{E}_{{s}_{t+1} \sim \txn} \Big [ \mathbb{E}_{{a}_{t+1} \sim \pi} [ Q_R({s}_{t+1}, {h}_{t+1}, {a}_{t+1}) \\
    & - \log \pi({a}_{t+1} | {s}_{t+1}, {h}_{t+1}) ] \Big ] \\[5pt]
    & Q_{\psi}(s_t, h_t, a_t) \\
    = & \log p({\psi}_{t} = 1 | {s}_{t}, {h}_{t}, {a}_{t}) + \\
    & \mathbb{E}_{{{s}_{t+1} \sim \txn}} \Big [ \mathbb{E}_{{a}_{t+1} \sim \pi} [ Q_{\psi}({s}_{t+1}, {h}_{t+1}, {a}_{t+1}) ] \Big ]
\end{aligned}
\end{equation}
with $h_{t+1}$ being the deterministic output of the safety model given $s_{t}, h_{t}, a_{t}$.

Substituting $Q_R$ and $Q_{\psi}$ into Equation~\eqref{eq:var_lb}, the maximization problem now only consists of the recursive functions and entropy term: 
\begin{equation} \label{eq:max_elbo}
\small
\begin{aligned} 
    \max_{\pi} \mathbb{E}_{({s}_{t}, {h}_{t}) \sim \hat{p}} \Bigg [ {}& \mathbb{E}_{{a}_{t} \sim \pi} \Big [ Q_R({s}_{t}, {h}_{t}, {a}_{t}) \\
    & + Q_{\psi}({s}_{t}, {h}_{t}, {a}_{t}) - \log \pi({a}_{t} | {s}_{t}, {h}_{t}) \Big ] \Bigg ]
\end{aligned}
\end{equation}
Equation~\eqref{eq:max_elbo} implicitly assumes that $Q_R$ and $Q_{\psi}$ are of the same scale. In practice, the two Q-values can be of vastly different scale and a scaling coefficient $\lambda$ is added as a coefficient of $Q_{\psi}$. Section~\ref{section:auto_lambda} further illustrates how this scaling coefficient $\lambda$ helps to adjust the degree of constraint compliance. Similarly, we add an entropy coefficient $\alpha$ to the entropy term to adjust the level of entropy required for sufficient exploration, same as the conventional SAC method~\cite{haarnoja2018soft}.

\subsubsection{$\safesach$ Algorithm} \label{section:safesach}

Recall that our policy $\pi$ is parameterized by $\theta$. To maximize the objective in~\eqref{eq:max_elbo}, we can we can compute the gradient and perform gradient ascent. Recall that $h_t$ depends only on $s_{t-1}, h_{t-1}, a_{t-1}$. We can compute the gradient by sampling from current policy since the following proportional relationship holds. 
\begin{equation} \label{eq:gradient_sample}
\small
\begin{aligned} 
    \nabla_{\theta} \mathbb{E}_{({s}_{t}, {h}_{t}) \sim \hat{p}} \Bigg [ {}& \mathbb{E}_{{a}_{t} \sim {\pi}_{\theta}} \Big [ Q_R({s}_{t}, {h}_{t}, {a}_{t}) \\[-10pt]
    &  \hspace{-30pt} \!+\! \lambda Q_{\psi}({s}_{t}, {h}_{t}, {a}_{t}) - \alpha \log \pi_{\theta}({a}_{t} | {s}_{t}, {h}_{t}) \Big ] \Bigg ] \\
    \propto \mathbb{E}_{({s}_{t}, {h}_{t}) \sim \hat{p}} \Bigg [ {}& \nabla_{\theta} 
    \mathbb{E}_{{a}_{t} \sim {\pi}_{\theta}} \Big [ Q_R({s}_{t}, {h}_{t}, {a}_{t}) \\[-10pt]
    & \hspace{-30pt} \!+\! \lambda Q_{\psi}({s}_{t}, {h}_{t}, {a}_{t}) - \alpha \log \pi_{\theta}({a}_{t} | {s}_{t}, {h}_{t}) \Big ] \Bigg ]
\end{aligned}
\end{equation}
We refer the reader to Appendix~\ref{section:PG_proof} for the proof of the above relationship. 

To obtain a low-variance gradient estimate, we employ the same reparameterization trick~\cite{haarnoja2018soft} to rewrite ${a}_{t}$ in Equation~\eqref{eq:gradient_sample}.
\begin{equation} \label{eq:reparam_elbo}
\small
\begin{aligned} 
    \mathbb{E}_{\substack{({s}_{t}, {h}_{t}) \sim \mathcal{D} \\ {\epsilon}_{t} \sim \mathcal{N}}} \Bigg [ {}& \mathbb{E}_{{a}_{t} \sim {\pi}_{\theta}} \Big [ \nabla_{\theta} Q_R({s}_{t}, {h}_{t}, f_{\theta}({\epsilon}_t; {s}_{t}, {h}_{t}) ) \\
    & + \lambda \nabla_{\theta}  Q_{\psi}({s}_{t}, {h}_{t}, f_{\theta}({\epsilon}_t; {s}_{t}, {h}_{t})) \\ 
    & - \alpha \nabla_{\theta} \log \pi(f_{\theta}({\epsilon}_t; {s}_{t}, {h}_{t}) | {s}_{t}, {h}_{t}) \Big ] \Bigg ]
\end{aligned}
\end{equation}
where ${\epsilon}_t$ is the noise vector sampled from the fixed distribution used in the policy (e.g. spherical Gaussian) and $\mathcal{D}$ is the replay buffer storing the experience tuple $\langle s, h, a, r, p_{\phi}({\psi} = 1 | s, h, a), s', h' \rangle$. We make $\safesach$ algorithm off-policy by sampling from replay buffer $\mathcal{D}$. Reusing the experience samples stored in replay buffer makes $\safesach$ more sample efficient. 

\subsubsection{Automatic Adjustment of $\lambda$ Coefficient} \label{section:auto_lambda}

As discussed in Section~\ref{section:var_inf}, a correct value of the scaling coefficient $\lambda$ helps to ensure constraint compliance. Using a fixed value for $\lambda$ can result in poor constraint compliance. Therefore, we propose a method to automatically learn the appropriate value of $\lambda$ during safe RL. 

Recall the non-Markovian CRL program in Equation~\eqref{eq:nmcrl}, the constraint consists of an exponent of sum. This is unwieldy and we need to rewrite it as a sum of log-probabilities to get a better estimate of its gradient. We use Jensen's inequality to derive the lower bound of this constraint:
\begin{equation} \label{eq:jensen_lb}
\small
\begin{aligned}
    {}& \mathbb{E}_{\tau \sim \pi_{\theta}} [ \exp(\sum_{t=0}^{T} \log {P}_{\phi}({\psi}_t = 1 | s_t, h_t, a_t)) ] \\
    & \geq \exp(\mathbb{E}_{\tau \sim \pi_{\theta}} [ \sum_{t=0}^{T} \log {P}_{\phi}({\psi}_t = 1 | s_t, h_t, a_t) ]) \\
    & \geq \exp(\log d)
\end{aligned}
\end{equation}
By requiring the lower bound function to be greater than $d$, it ensures that the original constraint is satisfied. We then re-express it on per timestep basis.
\begin{equation} \label{eq:constraint_dual}
\small
\begin{aligned}
& \hspace{-5pt} \mathbb{E}_{\tau \sim \pi_{\theta}} [ \frac{1}{T + 1} \sum_{t=0}^{T} \log {P}_{\phi}({\psi}_t = 1 | s_t, h_t, a_t) ] \geq \frac{\log d}{T + 1} \\
& \hspace{-5pt} \mathbb{E}_{\tau \sim \pi_{\theta}} \Big[ \mathbb{E}_{(s_t, h_t, a_t) \sim \tau} [ \log {P}_{\phi}({\psi}_t = 1 | s_t, h_t, a_t) ] \Big] \geq \frac{\log d}{T + 1} \\
& \hspace{-5pt} \mathbb{E}_{(s_t, h_t, a_t) \sim \rho_{\pi_{\theta}}} [ \log {P}_{\phi}({\psi} = 1 | s_t, h_t, a_t) ] \geq \frac{\log d}{T + 1}
\end{aligned}
\end{equation}
where $\rho_{\pi_{\theta}}$ refers to the distribution of $(s_t, h_t, a_t)$ by following policy ${\pi}_{\theta}$. 

With this rewritten constraint expression, the non-Markovian CRL program in Equation~\eqref{eq:nmcrl} becomes:
\begin{equation} \label{eq:nmcrl_rewritten}
\small
\begin{aligned}
 \max_{\theta} & \: \mathbb{E}_{(s_t, h_t, a_t) \sim \rho_{\pi_{\theta}}} [ \sum_{t=0}^{T} R(s_t, a_t) ] \\
  \textrm{s.t.} & \: \mathbb{E}_{(s_t, h_t, a_t) \sim \rho_{\pi_{\theta}}} [ \log {P}_{\phi}({\psi} = 1 | s, h, a) ] \geq \frac{\log d}{T + 1}
\end{aligned}
\end{equation}
We now write down its dual problem using Lagrangian method.
\begin{equation} \label{eq:dual_problem}
\small
\begin{aligned}
 {}& \min_{\lambda \geq 0} \max_{\theta} \: L(\lambda, \theta) \\
 = & \min_{\lambda \geq 0} \max_{\theta} \mathbb{E}_{(s_t, h_t, a_t) \sim \rho_{\pi_{\theta}}} [ \sum_{t=0}^{T} R(s_t, a_t) \\
 & + \lambda \log {P}_{\phi}({\psi}_t = 1 | s_t, h_t, a_t) - \lambda \frac{\log d}{T + 1} ] \\
\end{aligned}
\end{equation}
The inner maximization problem in~\eqref{eq:dual_problem}: maximization wrt $\theta$ is solved using $\safesach$ method in Equation~\eqref{eq:reparam_elbo}. The outer minimization problem (wrt to Lagrange Multiplier $\lambda$) can be iteratively solved by computing its gradient wrt $\lambda$.  
\begin{equation} \label{eq:optimal_lambda}
\small
\begin{aligned}
    {}& \nabla_{\lambda} L(\lambda, \theta) \\
    = & \mathbb{E}_{(s, h, a) \sim \rho_{\pi_{\theta}}} [\log P_{\phi}({\psi} = 1 | s, h, a) - \frac{\log d}{T + 1} )]
\end{aligned}
\end{equation}

\subsubsection{Practical Implementation} \label{section:practical_impl}

\textbf{Enriched Replay Buffer:} The experience tuples drawn from replay buffer $\mathcal{D}$ are $\langle s, h, a, r, p_{\phi}({\psi} = 1 | s, h, a), s', h' \rangle$. Once the safety model is trained, we can query it for the value of $h, p_{\phi}({\psi} = 1 | s, h, a), h'$, for a given conventional replay buffer experience tuple $\langle s, a, r, s' \rangle$. By enriching the existing replay buffer, we empower $\safesach$ to learn from past experiences and potentially improve its sample efficiency.

\textbf{Learning Safety Critic:} $\safesach$ learns two different critics: $Q_R$ as the regular reward critic and $Q_{\psi}$ as safety critic~\cite{ha2021learning}. Two separate deep learning networks are used to learn these two critics. 

\textbf{Learning $\lambda$:} Equation~\eqref{eq:optimal_lambda} relies on on-policy samples  $(s_i, h_i)$ to compute the gradient wrt $\lambda$. To approximate this, the $\safesach$ algorithm only draws recent $(s, h)$ pairs from the replay buffer to compute the gradient wrt $\lambda$.

We refer the readers to Appendix~\ref{appendix:algorithm} for the detailed pseudocode.

\section{Empirical Results}

The objective of our empirical experiments is to investigate: (1) Does $\safesach$ converge to a safe policy, respecting the non-Markovian constraint learned from labeled trajectories? (2) How quickly does $\safesach$ converge? (3) Does the extra information from hidden state $h$ and log-probability score $P(\psi = 1 | s, h, a)$ aid the policy in making informed decision? (4) What is the effect of using lower bound constraint in Equation~\eqref{eq:nmcrl_rewritten} for learning the optimal Lagrange multiplier $\lambda$? We designed the following experiment setup to support the investigation of the above research questions. 

\begin{figure*}[ht]
    \centering
    \includegraphics[width=\textwidth]{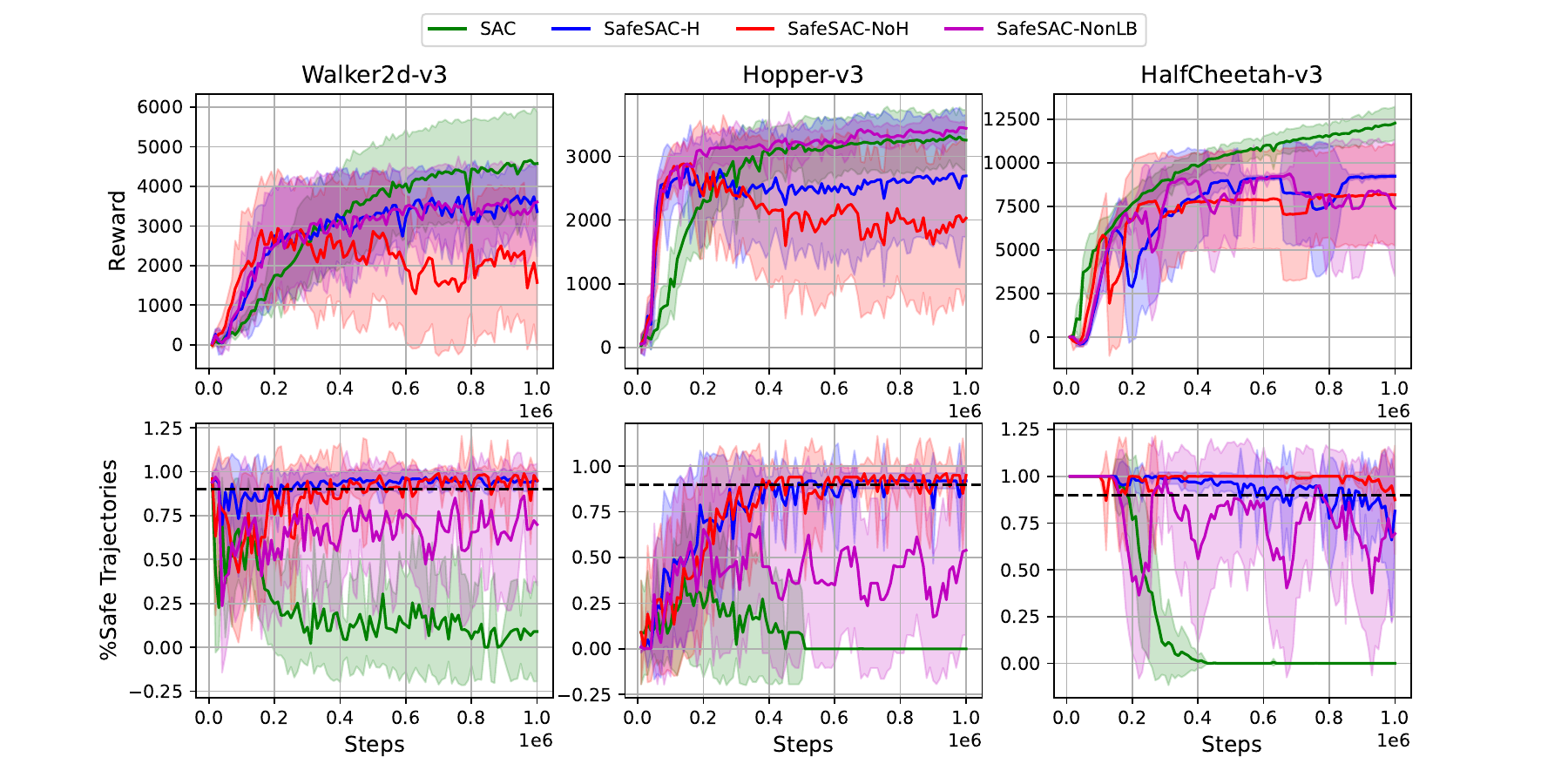}
    \vskip 0pt
    \caption{MuJoCo Experiments - Reward and Safety Performance }
    \label{fig:mujoco_short}
    \vskip 0pt
\end{figure*}
\subsection{Experiment Setup} \label{section:expt_setup}

\textbf{Tasks:} We experimented 9 control tasks in the following environments: 

(A) MuJoCo domain~\cite{todorov2012mujoco}: We consider three tasks: Walker2D, Hopper and HalfCheetah. All their reward objective is to move forward (on x-axis) as quickly as possible. We impose additional non-Markovian constraints on them. For Walker2D, the simple moving average of its resultant velocity for the last 50 timesteps must be lower than a set threshold. The non-Markovian constraint on Hopper specifies that the simple moving average of its z-velocity must be lower than a set threshold. Similarly for HalfCheetah, we constrain its moving average x-velocity. 

(B) Bullet Safety Gym~\cite{Gronauer2022BulletSafetyGym}: Our experiment involves two types of agents (i.e. Ball and Car) and each of them executes two different control tasks (i.e. Circle and Run). For Circle task, agent is rewarded by moving in circle with high angular velocity. We impose non-Markovian constraint where the agent must not leave the safety zone for more than 15\% of the time in the past 20 timesteps. The Run task requires the agent to run forward as fast as possible. We designed a non-Markovian constraint which disallows the agent to run faster than a set velocity or leave the safety zone more than 15\% of the time in the past 20 timesteps. 

(C) RDDL Gym~\cite{taitler2022pyrddlgym}: We experiment two domains: Navigation and Air-Conditioning tasks. We refer the readers to Appendix~\ref{appendix:rddl_gym} for the detailed experiment settings and results. 

We highlight that all the non-Markovian constraints are not known to the safety model or RL agent. The safety model only learns the constraint from labeled trajectories.

\textbf{Safety Model:} To train our safety model, we collect the trajectories traversed by an unconstrained SAC agent during its learning. For each task, we collect about 400K trajectories. A python script is used to compute the safety label for these trajectories. The dataset is also further augmented by labeling sub-trajectories contained within these 400K trajectories. The final classification accuracy of our trained safety model is around 97-99\% for the tasks mentioned.

\textbf{Baseline:} To the best of our knowledge, existing safe RL algorithms are unable to handle unknown non-Markovian constraints. Therefore, we design the following baselines in our result reporting. (1) \textit{SAC}: Our method $\safesach$ is compared against unconstrained SAC to assess its constraint compliance and speed to convergence; (2) \textit{SafeSAC-NoH}: The purpose of comparing against this baseline is to inspect the effect of hidden state $h$. It has the same algorithm as $\safesach$ but its policy $\pi(a | s)$ and critics $Q_R(s, a), Q_{\psi}(s, a)$ do not depend on $h$. The same log-probability score $\log P_{\phi}(\psi = 1 | s, h, a)$ is used for the learning of safety critic $Q_{\psi}(s, a)$, with $h$ being marginalized out for a given state-action pair
; (3) \textit{SafeSAC-NonLB}: This algorithm is almost identical to $\safesach$ but it learns Lagrange multiplier $\lambda$ by computing gradient of the original constraint expression (i.e. LHS of Equation~\eqref{eq:jensen_lb}). This baseline reveals the effectiveness of our derived gradient in Equation~\eqref{eq:optimal_lambda}. 

\textbf{Metrics:} Following the original non-Markovian CRL program defined in Equation~\eqref{eq:nmcrl}, we assess the performance mainly using two metrics: total return and percentage of trajectories respecting non-Markovian safety constraints. We highlight that the safety metric we report is the percentage of trajectories violating the ground-truth non-Markovian constraint defined in Section~\ref{section:expt_setup}, not the predicted probability score obtained from safety model. This is a more stringent metric since a safety model can never perfectly predict the probability of safe / unsafe behaviors all the time. Appendix~\ref{section:add_results} contains an additonal metric for MuJoCo and Bullet Gym environment: percentage of steps where the window average requirement is met.

\textbf{Performance Reporting:} We report the metrics by evaluating 100 separate test trajectories after a fixed number of learning steps. These trajectories are generated using a different evaluation environment, separate from the training environment. The metrics are reported in the terms of average and standard deviation values across 10 training seeds. 
\begin{figure*}[ht]
    \centering
    \includegraphics[width=\textwidth]{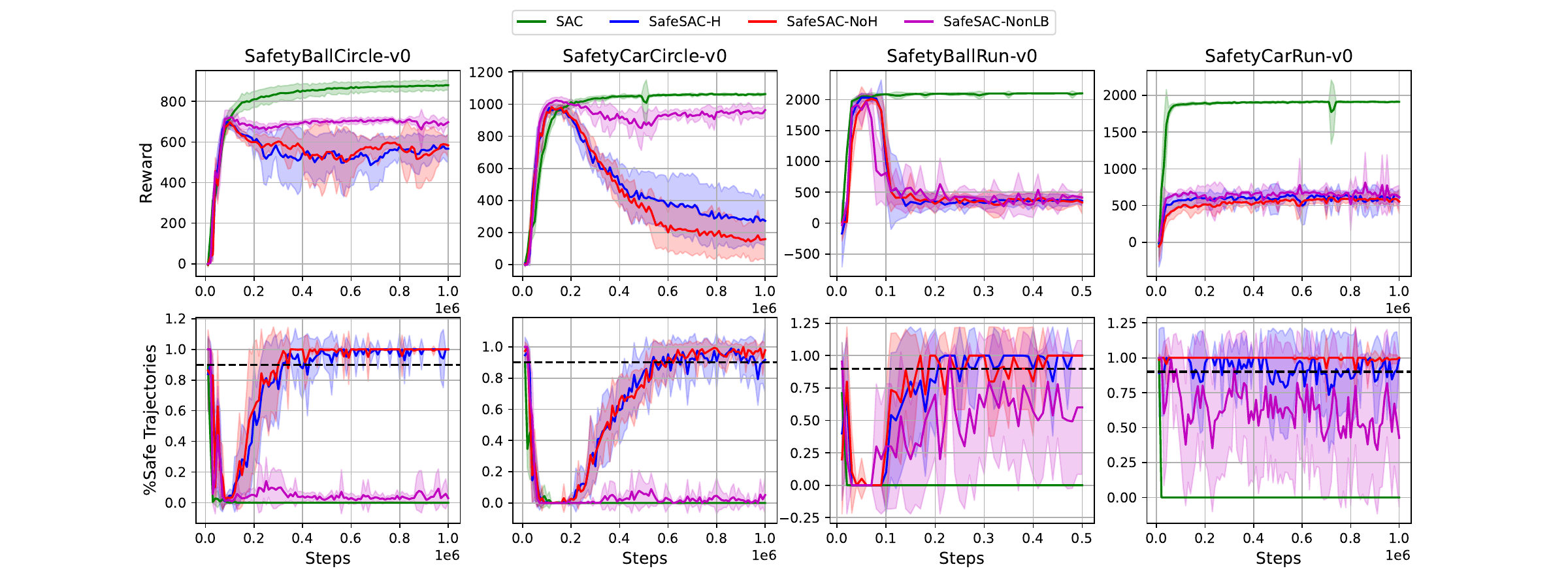}
    \vskip 0pt
    \caption{Safety Bullet Gym Experiments - Reward and Safety Performance }
    \label{fig:bulletgym_short}
    \vskip 0pt
\end{figure*}

\subsection{Main Results and Analysis}

Figure~\ref{fig:mujoco_short} and Figure~\ref{fig:bulletgym_short} depict the experiment result for MuJoCo and Bullet Safety Gym tasks respectively. The black dashed line in the second row indicates our safety target: 90\% of the generated trajectories comply with the non-Markovian constraint throughout their entire time-length.

From both set of figures, it can be observed that $\safesach$ converges to safe yet good quality solution while SAC quickly converges to unsafe solution even though it attains higher return. This demonstrates that $\safesach$ is able to learn a policy reasonably quickly while respecting the non-Markovian safety constraint.

It is worth highlighting that in Walker2D and Hopper, $\safesach$ achieves a higher reward than SAC in the early part of training (below 200K steps). This is due to the ability of $\safesach$ in learning from experiences stored in enriched replay buffer. Being an off-policy algorithm, $\safesach$ is capable of reusing data (previously used for the training of safety model) and facilitate safe RL in a sample-efficient manner. 

\subsubsection{Importance of Hidden State $h$} \label{section:h_ablation}

To evaluate the effect of having $h$ in the policy and critics, we compare the performance of $\safesach$ against SafeSAC-NoH. It can be observed that $\safesach$ outperforms SafeSAC-NoH in Walker, Hopper and Car-Circle tasks. 

The summary vector $h$ provides the safety critic $Q_{\psi}(s, h, a)$ an additional ability to differentiate between safe and unsafe actions when a history pattern is observed. As a result, its policy $\pi_{\theta}(a | s, h)$ is also more nuanced since it can decide if it needs to be extra careful when a history pattern is observed through $h$. In contrast, SafeSAC-NoH has no awareness of the past history since its critic and policy does not depend on $h$. The safety critic $Q_{\psi}(s, a)$ just learns a single scalar value for all the possible $h$ permutations that might appear with this state-action pair. Consequently, it is not able to facilitate safe RL as well as $\safesach$. 

Interestingly, despite being inferior to $\safesach$, SafeSAC-NoH still performs decently in almost all tasks. This is because SafeSAC-NoH also trains its critic based on the log-probability scores $\log P_{\phi}(\psi = 1 | s, h, a)$ from safety model. This shows that our safety model is still useful for the conventional critic $Q_{\psi}(s, a)$ and policy $\pi_{\theta}(a | s)$. They can potentially learn the association between safe / unsafe history and current state, and subsequently exploit this indirect association to facilitate non-Markovian safe RL.

\subsubsection{Automatic Adjustment of $\lambda$}

It is crucial for $\safesach$ to compute the correct value of Lagrange multiplier $\lambda$ in each learning iteration. This value provides key information about the correct tradeoff between reward maximization and constraint satisfaction, helping $\safesach$ satisfy constraint threshold. To evaluate how well $\safesach$ adjusts this tradeoff, we compare it against SafeSAC-NonLB. 

The most striking observation in Figure~\ref{fig:mujoco_short} and~\ref{fig:bulletgym_short} is that SafeSAC-NonLB fails to satisfy the constraint threshold in all tasks. This implies that SafeSAC-NonLB underestimates the correct value of $\lambda$. Consequently, SafeSAC-NonLB agent overweights reward maximization and fails to converge to feasible policy. In contrast, $\safesach$ estimates $\lambda$ better and successfully balances between reward maximization and constraint satisfaction. 

\section{Conclusion}

In this work, we study non-Markovian safety constraint and demonstrate using a safety model to learn such constraint and facilitate safe RL. We propose a unique design of the safety model and employ variational inference method to reformulate non-Markovian CRL program and derive our safe RL algorithm $\safesach$. Our proposed method uses the log-probability scores from the safety model to train a safety critic and policy. These deep learning models form the key components of our proposed algorithm. Unlike previous safe RL algorithms which focus on Markovian safety cost function, $\safesach$ solves for a safe policy which respect non-Markovian safety constraints inferred from labeled trajectory data. 

\section*{Impact Statement}

This paper presents work whose goal is to advance the field of Machine Learning. There are many potential societal consequences of our work, none which we feel must be specifically highlighted here.


\bibliography{SafeSACH}
\bibliographystyle{icml2024}

\newpage
\appendix
\onecolumn
\section{Proofs} \label{appendix:proofs}

\subsection{Optimal Distribution of Trajectory $p(\tau | \boldsymbol{o}_{0:T}, \boldsymbol{\psi}_{0:T})$} \label{appendix:proof_p_tau}

Conditioning on \(\mathcal{O}_t = 1, {\psi}_t = 1\) for all \(t \in \{0, ..., T\}\), the posterior distribution of observing an optimal trajectory $\tau$ is: 
\begin{equation} \nonumber
\begin{aligned}
 & p(\tau | \boldsymbol{o}_{0:T}, \boldsymbol{\psi}_{0:T}) \propto p(\tau, \boldsymbol{o}_{0:T}, \boldsymbol{\psi}_{0:T}) \\
 = & \int_{{h}_{0:T}} p({s}_0)p({h}_0) \prod_{t=0}^{T-1} [p({a}_t) p({s}_{t+1} | {s}_{t}, {a}_{t}) p(\mathcal{O}_t = 1 | {s}_{t}, {a}_{t}) p({h}_{t+1} | {s}_{t}, {h}_{t}, {a}_{t}) p({\psi}_t = 1 | {s}_{t}, {h}_{t}, {a}_{t})] p({a}_T) \\
 & p(\mathcal{O}_T = 1 | {s}_{T}, {a}_{T}) p({\psi}_T = 1 | {s}_{T}, {h}_{T}, {a}_{T}) \, dh_{0:T}
\end{aligned}
\end{equation}

Recall from Section~\ref{section:safety_model} that \({h}_0\) is a fixed all-zero vector and \({h}_{t+1}\) is deterministically determined given \(({s}_{t}, {h}_{t}, {a}_{t})\). Thus, the posterior distribution reduces to: 
\begin{equation} \label{eq:jointP}
\begin{aligned}
& p(\tau, \boldsymbol{o}_{0:T}, \boldsymbol{\psi}_{0:T}) \\
= & p({s}_0) \prod_{t=0}^{T-1} [ p({a}_t) p({s}_{t+1} | {s}_{t}, {a}_{t}) p(\mathcal{O}_t = 1 | {s}_{t}, {a}_{t}) p({\psi}_t = 1 | {s}_{t}, {h}_{t}, {a}_{t}) ] p({a}_T) p(\mathcal{O}_T = 1 | {s}_{T}, {a}_{T}) \\
 & p({\psi}_T = 1 | {s}_{T}, {h}_{T}, {a}_{T}) \\
 = & p({s}_0) \prod_{t=0}^{T-1} [p({a}_t) p({s}_{t+1} | {s}_{t}, {a}_{t}) \exp(r({s}_t, {a}_t)) \exp(\log p({\psi}_t = 1 | {s}_{t}, {h}_{t}, {a}_{t}))]  p({a}_T) p\exp(r({s}_T, {a}_T)) \\
 & \exp(\log p({\psi}_T = 1 | {s}_{T}, {h}_{T}, {a}_{T}))]\\
 = & \left[p({s}_0) \prod_{t=0}^{T-1} p({a}_t) p({s}_{t+1} | {s}_{t}, {a}_{t}) \right] p(a_T) \exp(\sum_{t=0}^{T} r({s}_t, {a}_t) + \log p({\psi}_t = 1 | {s}_{t}, {h}_{t}, {a}_{t}))  \\
\end{aligned}
\end{equation}

The term $p({a}_t)$ is further eliminated from the equation by considering uniform action prior $p(a_t)$ assumption~\cite{levine2018reinforcement} and the posterior distribution further reduces to:
\begin{equation} \label{eq:p_tau}
    \begin{aligned}
        p(\tau, \boldsymbol{o}_{0:T}, \boldsymbol{\psi}_{0:T}) = p({s}_0) \prod_{t=0}^{T-1} [p({s}_{t+1} | {s}_{t}, {a}_{t}) \exp(r({s}_t, {a}_t)) p({\psi}_t = 1 | {s}_{t}, {h}_{t}, {a}_{t})] \exp(r({s}_T, {a}_T)) p({\psi}_T = 1 | {s}_{T}, {h}_{T}, {a}_{T})
    \end{aligned}
\end{equation}

\subsection{KL-Divergence between Optimal and Approximate Trajectory Distributions} \label{appendix:kl_divergence}

With the optimal distribution $p(\tau, \boldsymbol{o}_{0:T}, \boldsymbol{\psi}_{0:T})$ expressed in Equation~\eqref{eq:optimal_tau} and approximate distribution $\hat{p}(\tau; \pi)$ expressed in Equation~\eqref{eq:approx_tau}, we can compute the KL-Divergence between the two distributions:

\begin{equation} \label{eq:kl_divergence_derive}
    \begin{aligned}
        {}& -D_{KL}(\hat{p}(\tau; \pi) || p(\tau, \boldsymbol{o}_{0:T}, \boldsymbol{\psi}_{0:T})) \\
        = & E_{\tau \sim \hat{p}(\tau)} \left[ \sum_{t=0}^T r({s}_t, {a}_t) + \log p({\psi}_t = 1 | {s}_{t}, {h}_{t}, {a}_{t}) -  \log \pi({a_t} | {s_t}, {h_t}) \right] \\
        = & \sum_{t=0}^T E_{({s}_{t}, {a}_{t}) \sim \hat{p}({s}_{t}, {a}_{t})} [r({s}_t, {a}_t)] + E_{({s}_{t}, {h}_{t}, {a}_{t}) \sim \hat{p}({s}_{0:t}, {a}_{0:t})} [\log p({\psi}_t = 1 | {s}_{t}, {h}_{t}, {a}_{t})] \\
        & + E_{({s}_{t}, {h}_{t}) \sim \hat{p}({s}_{0:t}, {a}_{0:t})} [\mathcal{H}(\pi({a_t} | {s_t}, {h_t}))] \\
        = & \mathbb{E}_{(s_{0:T}, a_{0:T}) \sim \hat{p}(s_{0:T}, a_{0:T})} \Big [ \sum_{t=0}^{T} r({s}_t, {a}_t) + \log p({\psi}_t = 1 | {s}_{t}, {h}_{t}, {a}_{t}) - \log \pi({a_t} | {s_t}, {h_t}) \Big ]
    \end{aligned}
\end{equation}

\subsection{Applicability of Policy Gradient Theorem in $\safesach$} \label{section:PG_proof}

Our objective is $J(\theta) = \sum_{s, h \in \mathcal{S}, \mathcal{H}} d^{\pi}(s, h) V^{\pi}(s, h)$. Let's first consider value function for a $(s, h)$ pair: 
\begin{equation} \label{eq:delta_V_sh_part1}
    \begin{split} 
        & \nabla_{\theta} V^{\pi}(s, h) \\
        & = \nabla_{\theta} \Big ( \mathbb{E}_{a \sim \pi_{\theta}} [Q^{\pi}_{R}(s, h, a) + Q^{\pi}_{\psi}(s, h, a) - \log \pi_{\theta}(a | s, h)] \Big ) \\
        & = \nabla_{\theta} \Big ( \int_{a} \pi_{\theta}(a | s, h) [Q^{\pi}_{R}(s, h, a) + Q^{\pi}_{\psi}(s, h, a) - \log \pi_{\theta}(a | s, h)] \, da \Big ) \\
        & = \int_{a} \Big ( \nabla_{\theta} \pi_{\theta}(a | s, h) (Q^{\pi}_{R}(s, h, a) + Q^{\pi}_{\psi}(s, h, a)) + \pi_{\theta}(a | s, h) \nabla_{\theta} (Q^{\pi}_{R}(s, h, a) + Q^{\pi}_{\psi}(s, h, a)) \\
        & \qquad \qquad - \log \pi_{\theta}(a | s, h) \nabla_{\theta} \pi_{\theta}(a | s, h) - \nabla_{\theta} \pi_{\theta}(a | s, h) \Big ) \, da \\
        & = \int_{a} \Bigg [ \nabla_{\theta} \pi_{\theta}(a | s, h) (Q^{\pi}_{R}(s, h, a) + Q^{\pi}_{\psi}(s, h, a)) \\
        & \qquad \qquad + \pi_{\theta}(a | s, h) \, \nabla_{\theta} \Big ( r(s, a) + \log p(\psi = 1 | s, h, a) + \int_{s', h'} p(s', h' | s, h, a) V^{\pi}(s', h') \, ds' \, dh' \Big ) \\
        & \qquad \qquad - (\log \pi_{\theta}(a | s, h) + 1) \, \nabla_{\theta} \pi_{\theta}(a | s, h) \Bigg ] \, da \\
        & = \int_{a} \Bigg [ \nabla_{\theta} \pi_{\theta}(a | s, h) (Q^{\pi}_{R}(s, h, a) + Q^{\pi}_{\psi}(s, h, a)) - (\log \pi_{\theta}(a | s, h) + 1) \, \nabla_{\theta} \pi_{\theta}(a | s, h) \\
        & \qquad \qquad + \pi_{\theta}(a | s, h) \int_{s', h'} p(s', h' | s, h, a) \nabla_{\theta} V^{\pi}(s', h') \, ds' \, dh' \Bigg ] \, da \\
        & = \int_{a} \Bigg [ \nabla_{\theta} \pi_{\theta}(a | s, h) (Q^{\pi}_{R}(s, h, a) + Q^{\pi}_{\psi}(s, h, a)) + \pi_{\theta}(a | s, h) \, \mathbb{E}_{s', h'} \Big [ \nabla_{\theta} V^{\pi}(s', h') \Big ] \\
        & \qquad \qquad - (\log \pi_{\theta}(a | s, h) + 1) \, \nabla_{\theta} \pi_{\theta}(a | s, h) \Bigg ] \, da \\
        & = \int_{a} \Bigg [ \nabla_{\theta} \pi_{\theta}(a | s, h) \Big ( Q^{\pi}_{R}(s, h, a) + Q^{\pi}_{\psi}(s, h, a) - \log \pi_{\theta}(a | s, h) - 1 \Big ) + \pi_{\theta}(a | s, h) \mathbb{E}_{s', h'} \Big [ \nabla_{\theta} V^{\pi}(s', h') \Big ] \Bigg ] \, da
    \end{split}
\end{equation}
Let $\phi(s, h) = \int_{a} \nabla_{\theta} \pi_{\theta}(a | s, h) \Big ( Q^{\pi}_{R}(s, h, a) + Q^{\pi}_{\psi}(s, h, a) - \log \pi_{\theta}(a | s, h) - 1 \Big ) \, da$, $\nabla_{\theta} V^{\pi}(s, h)$ is then:
\begin{equation} \label{eq:delta_V_sh_part2}
    \begin{split}
        & \nabla_{\theta} V^{\pi}(s, h) \\
        & = \phi(s, h) + \int_{a} \pi_{\theta}(a | s, h) \mathbb{E}_{s', h'} \Big [ \nabla_{\theta} V^{\pi}(s', h') \Big ] \, da \\
        & = \phi(s, h) + \int_{a, s', h'} \pi_{\theta}(a | s, h) p(s', h' | s, h, a) \, \nabla_{\theta} V^{\pi}(s', h') \, da \, ds' \, dh' \\
        & = \phi(s, h) + \int_{s', h'} {\rho}^{\pi}((s, h) \rightarrow (s', h'), 1) \nabla_{\theta} V^{\pi}(s', h') \, ds' \, dh' \\
        & = \phi(s, h) + \int_{s', h'} {\rho}^{\pi}((s, h) \rightarrow (s', h'), 1) \Big [ \phi(s', h') \\
        & \qquad \qquad \qquad \qquad + \int_{s'', h''} {\rho}^{\pi}((s', h') \rightarrow (s'', h''), 1) \nabla_{\theta} V^{\pi}(s'', h'') \, ds'' \, dh'' \Big ] \, ds' \, dh' \\
        & =  \phi(s, h) + \int_{s', h'} {\rho}^{\pi}((s, h) \rightarrow (s', h'), 1) \phi(s', h') \, ds' \, dh' \\
        & \qquad \qquad \qquad \qquad + \int_{s'', h''} {\rho}^{\pi}((s, h) \rightarrow (s'', h''), 2) \nabla_{\theta} V^{\pi}(s'', h'') \, ds'' \, dh'' \\
        & =  \phi(s, h) + \int_{s', h'} {\rho}^{\pi}((s, h) \rightarrow (s', h'), 1) \phi(s', h') \, ds' \, dh' \\
        & \qquad \qquad \qquad \qquad + \int_{s'', h''} {\rho}^{\pi}((s, h) \rightarrow (s'', h''), 2) \phi(s'', h'') \, ds'' \, dh'' \\
        & \qquad \qquad \qquad \qquad + \int_{s''', h'''} {\rho}^{\pi}((s, h) \rightarrow (s''', h'''), 3) \nabla_{\theta} V^{\pi}(s''', h''') \, ds''' \, dh''' \\
        & = ... \\
        & = \int_{s', h'} \sum_{k=0}^{\infty} {\rho}^{\pi}((s, h) \rightarrow (s', h'), k) \, \phi(s', h') \, ds' \, dh'
    \end{split}
\end{equation}

Next, we let $\eta(s, h) = \sum_{k=0}^{\infty} {\rho}^{\pi}((s_0, h_0) \rightarrow (s, h), k)$. The gradient of the objective $\nabla_{\theta} J(\theta)$ is re-expressed as: 

\begin{equation} \label{eq:delta_J_theta_part1}
    \begin{split}         
        & \nabla_{\theta} J(\theta) \\
        & = \nabla_{\theta}  V^{\pi}(s_0, h_0) \\
        & = \int_{s, h} \sum_{k=0}^{\infty} {\rho}^{\pi}((s_0, h_0) \rightarrow (s, h), k) \, \phi(s, h) \, ds \, dh \\
        & = \int_{s, h} \eta(s, h) \, \phi(s, h) \, ds \, dh \\
        & = \Big ( \int_{s', h'} \eta(s', h') \, ds' \, dh' \Big ) \int_{s, h} \frac{\eta(s, h)}{\int_{s', h'} \eta(s', h') \, ds' \, dh'} \, \phi(s, h) \, ds \, dh \\
        & \propto \int_{s, h} \frac{\eta(s, h)}{\int_{s', h'} \eta(s', h') \, ds' \, dh'} \, \phi(s, h) \, ds \, dh
    \end{split}
\end{equation}

Finally, let $d^{\pi}(s, h) = \frac{\eta(s, h)}{\int_{s', h'} \eta(s', h') \, ds' \, dh'}$ to further simplify the gradient $\nabla_{\theta} J(\theta)$.

\begin{equation} \label{eq:delta_J_theta_part2}
    \begin{split}         
        & \nabla_{\theta} J(\theta) \\
        & = \int_{s, h} d^{\pi}(s, h) \, \phi(s, h) \, ds \, dh \\
        & = \int_{s, h} d^{\pi}(s, h) \int_{a} \nabla_{\theta} \pi_{\theta}(a | s, h) \Big ( Q^{\pi}_{R}(s, h, a) + Q^{\pi}_{\psi}(s, h, a) - \log \pi_{\theta}(a | s, h) - 1 \Big ) \, da \, ds \, dh \\
        & = \int_{s, h} d^{\pi}(s, h) \nabla_{\theta} \int_{a}  \pi_{\theta}(a | s, h) \Big ( Q^{\pi}_{R}(s, h, a) + Q^{\pi}_{\psi}(s, h, a) - \log \pi_{\theta}(a | s, h) - 1 \Big ) \, da \, ds \, dh \\
        & = \mathbb{E}_{(s, h) \sim d^{\pi}} \Bigg [ \nabla_{\theta} \mathbb{E}_{a \sim \pi} \Big [ \Big ( Q^{\pi}_{R}(s, h, a) + Q^{\pi}_{\psi}(s, h, a) - \log \pi_{\theta}(a | s, h) - 1 \Big ) \Big ] \Bigg ]
    \end{split}
\end{equation}

\section{Algorithm} \label{appendix:algorithm}

\begin{algorithm}[thb]
\caption{$\safesach$ algorithm}
\label{algS:algorithm}
\textbf{Models}: ${P}_{\phi}({\psi} = 1 | s, h, a)$, $Q_R(s, h, a | \omega)$, $Q_{\psi}(s, h, a | \nu)$ and $\pi_{\theta}(a | s, h)$
\begin{algorithmic}[1] 
\small
\STATE Randomly initialize weights $\phi$, $\omega$, $\nu$, and\, $\theta$
\STATE Train safety model $\phi$ using labeled trajectory data
\STATE Enrich existing replay buffer $\mathcal{D}$ by querying safety model for the corresponding $(h, P_{\phi}({\psi} = 1 | s, h, a), h')$
\STATE \textbf{repeat} until convergence
\STATE \qquad Observe state $s$ and $h$, execute action $a \sim \pi_{\theta}(\cdot | s, h)$
\STATE \qquad Observer next state $s'$, reward $r$, and done signal $d$
\STATE \qquad Query safety model for $h'$ and $P_{\phi}({\psi} = 1 | s, h, a)$
\STATE \qquad Store $(s, h, a, r, P_{\phi}({\psi} = 1 | s, h, a), s', h', d)$ in replay buffer $\mathcal{D}$
\STATE \qquad If $s'$ is terminal, reset environment state
\STATE \qquad \textbf{if} it is time to update \textbf{then}
\STATE \qquad \qquad Draw recent state-action pairs $(s, h)$ from $\mathcal{D}$, query action $a \sim \pi_{\theta}(\cdot | s, h)$ and update $\lambda$ by computing~\eqref{eq:optimal_lambda}
\STATE \qquad \qquad \textbf{for} $j$ in range(number of updates) \textbf{do}
\STATE \qquad \qquad \qquad Randomly sample batch of transitions $B={(s, h, a, r, P_{\phi}({\psi} = 1 | s, h, a), s', h', d)}$ from $\mathcal{D}$
\STATE \qquad \qquad \qquad Update targets for the $Q_R$ functions:
\begin{equation}
\small
\begin{aligned}
    & y_R(r, s', h', d) = r + \gamma (1 - d) \Big ( \min_{i=1,2} Q_{R_{targ}, i}(s', h', \Tilde{a}') - \alpha \log \pi_{\theta}(\Tilde{a}' | s', h') \Big ), \Tilde{a}' \sim \pi_{\theta}(\cdot | s', h')
\end{aligned}
\end{equation}
\STATE \qquad \qquad \qquad Update $Q_R$ by one step gradient descent using:
\begin{equation}
\small
\begin{aligned}
    & \nabla_{\omega_i} \frac{1}{|B|} \sum_{(s, h, a, r, s', h', d) \in B} (Q_{R_i}(s, h, a) - y_R(r, s', h', d))^2 \qquad \text{ for } i = 1, 2
\end{aligned}
\end{equation}
\STATE \qquad \qquad \qquad Update targets for the $Q_{\psi}$ functions:
\begin{equation}
\small
\begin{aligned}
    & y_{\psi}(s, h, a, s', h', d) = P_{\phi}({\psi} = 1 | s, h, a) + \gamma (1 - d) \Big ( \min_{i=1,2} Q_{{\psi}_{targ}, i}(s', h', \Tilde{a}') \Big ), \Tilde{a}' \sim \pi_{\theta}(\cdot | s', h')
\end{aligned}
\end{equation}
\STATE \qquad \qquad \qquad Update $Q_{\psi}$ by one step gradient descent using:
\begin{equation}
\small
\begin{aligned}
    & \nabla_{\nu_i} \frac{1}{|B|} \sum_{(s, h, a, r, s', h', d) \in B} (Q_{{\psi}_i}(s, h, a) - y_{\psi}(s, h, a, s', h', d))^2 \qquad \text{ for } i = 1, 2
\end{aligned}
\end{equation}
\STATE \qquad \qquad \qquad Update policy by one step gradient descent using:
\begin{equation}
\small
\begin{aligned}
    & \nabla_{\theta} \frac{1}{|B|} \sum_{(s, h) \in B} \Big ( \min_{i=1,2} Q_{R_i}(s, h, \Tilde{a}_{\theta}(s, h)) + \lambda \min_{i=1,2} Q_{{\psi}_i}(s, h, \Tilde{a}_{\theta}(s, h)) - \alpha \log \pi_{\theta}(\Tilde{a}_{\theta}(s, h) | s, h) \Big ) \\
    & \text{where } \Tilde{a}_{\theta}(s, h) \text{ is a sample from }\pi_{\theta}(\cdot | s', h') \text{ which is differentiable wrt } \theta \text{ via reparameterization trick}
\end{aligned}
\end{equation}
\STATE \qquad \qquad \qquad Update target networks by polyak averaging
\end{algorithmic}
\end{algorithm}

\section{Code Appendix} \label{appendix:code}

We provide a sample code snippet of our $\safesach$ implementation in the supplementary materials. If our paper is accepted, we will release the codes to be hosted on an online code repository (i.e. GitHub). 

\section{Experiment Settings} \label{section:expt_settings}

\subsection{Deep Learning Architecture} \label{appendix:nw_arch}

\begin{table}[th]
\begin{center}
\begin{tabular}{ | c | c | c | c | c | c |  }
\hline
\textbf{Neural Network Model} & \textbf{Hidden Units} & \textbf{Normalization} & \textbf{Learning Rate} & \textbf{Minibatch Size} & \textbf{Others} \\
\hline
Classifier & [32, 32] & Batch Norm & 0.0001 & 100 & Dropout: 0.5 \\
Policy Net & [32, 32] & Layer Norm & 0.0003 & 100 & PPO Clip: 0.2 \\
Value Net & [32, 32] & Layer Norm & 0.0003 & 100 & - \\
\hline
\end{tabular}
\caption{Hyperparameter Settings of Neural Network Models for Grid World Experiments}
\label{tableS:hyperparam}
\end{center}
\end{table}


\section{Additional Empirical Experiments} \label{section:add_results}

\subsection{RDDL Gym Description} \label{appendix:rddl_gym}

RDDL Gym~\cite{taitler2022pyrddlgym}: We experiment two domains: Navigation and Air-Conditioning tasks. In Navigation task, agent is rewarded by moving towards the goal. We designated a dangerous zone and the agent must not stay in the dangerous zone more than 1 timestep in the entire trajectory. For Air-Conditioning task, reward is given by maintaining the connected rooms at desired temperature range. The non-Markovian constraint imposed is that the server room must not have temperature higher than a preset threshold more than or equal to two consecutive timesteps.

\subsection{Additional Results}
\begin{figure*}[ht]
    \centering
    \includegraphics[width=0.7\textwidth]{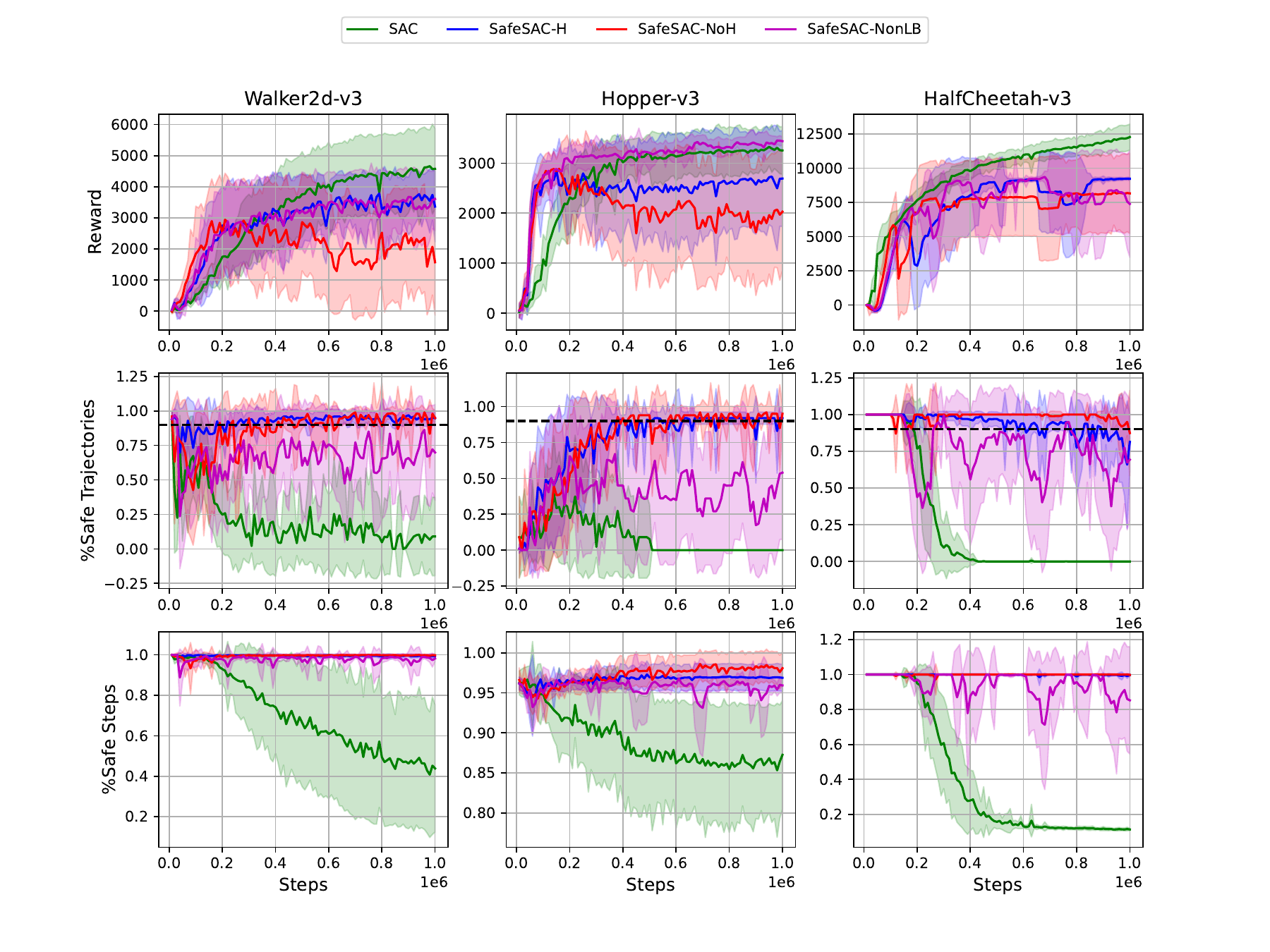}
    \vskip 0pt
    \caption{MuJoCo Results - Reward and Safety Performance}
    \label{fig:mujoco_full}
    \vskip 0pt
\end{figure*}

\begin{figure*}[htb]
    \centering
    \includegraphics[width=0.7\textwidth]{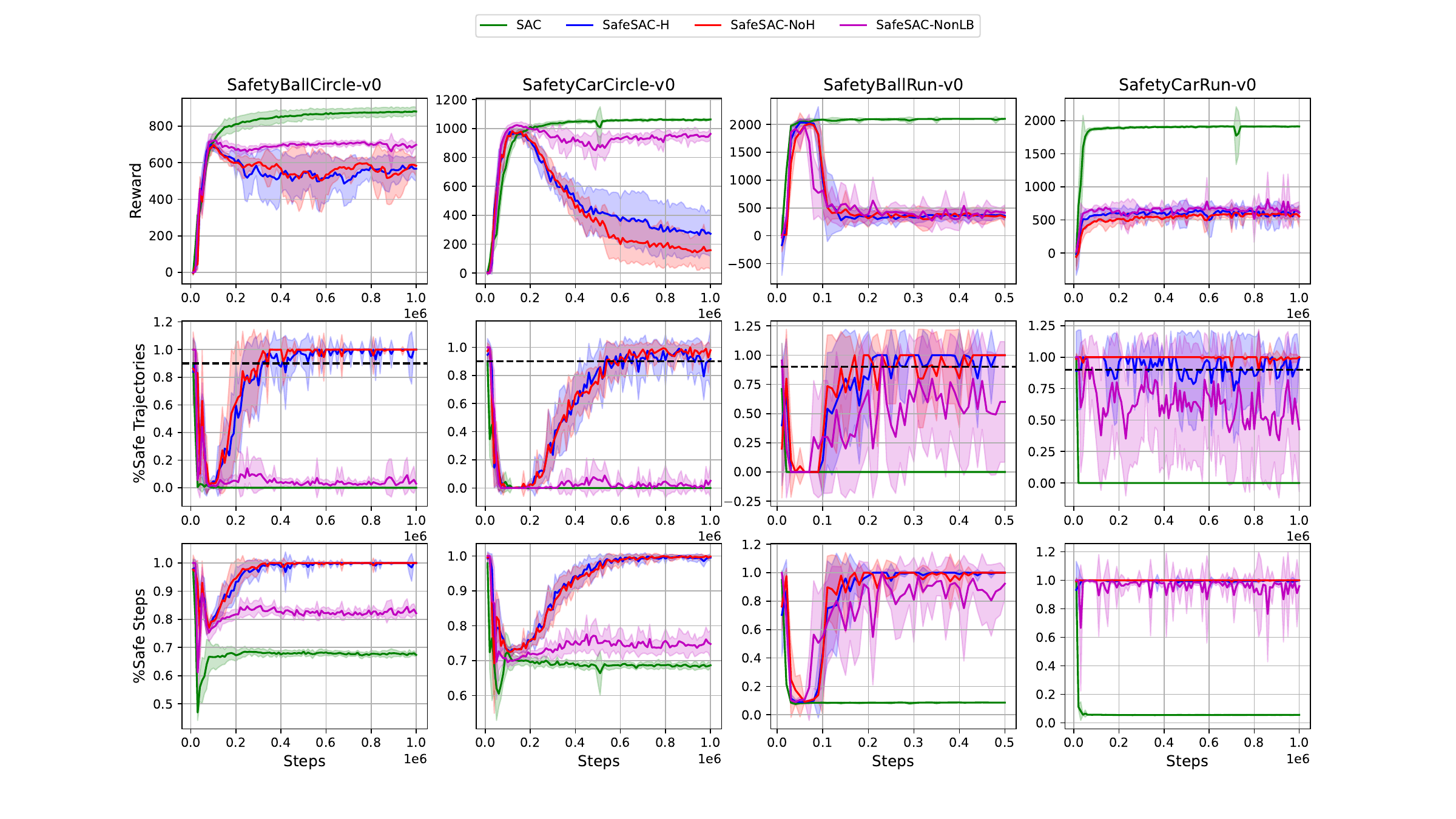}
    \vskip 0pt
    \caption{Bullet Safety Gym Results - Reward and Safety Performance}
    \label{fig:bulletgym_full}
    \vskip 0pt
\end{figure*}

\begin{figure*}[htb]
    \centering
    \includegraphics[width=0.7\textwidth]{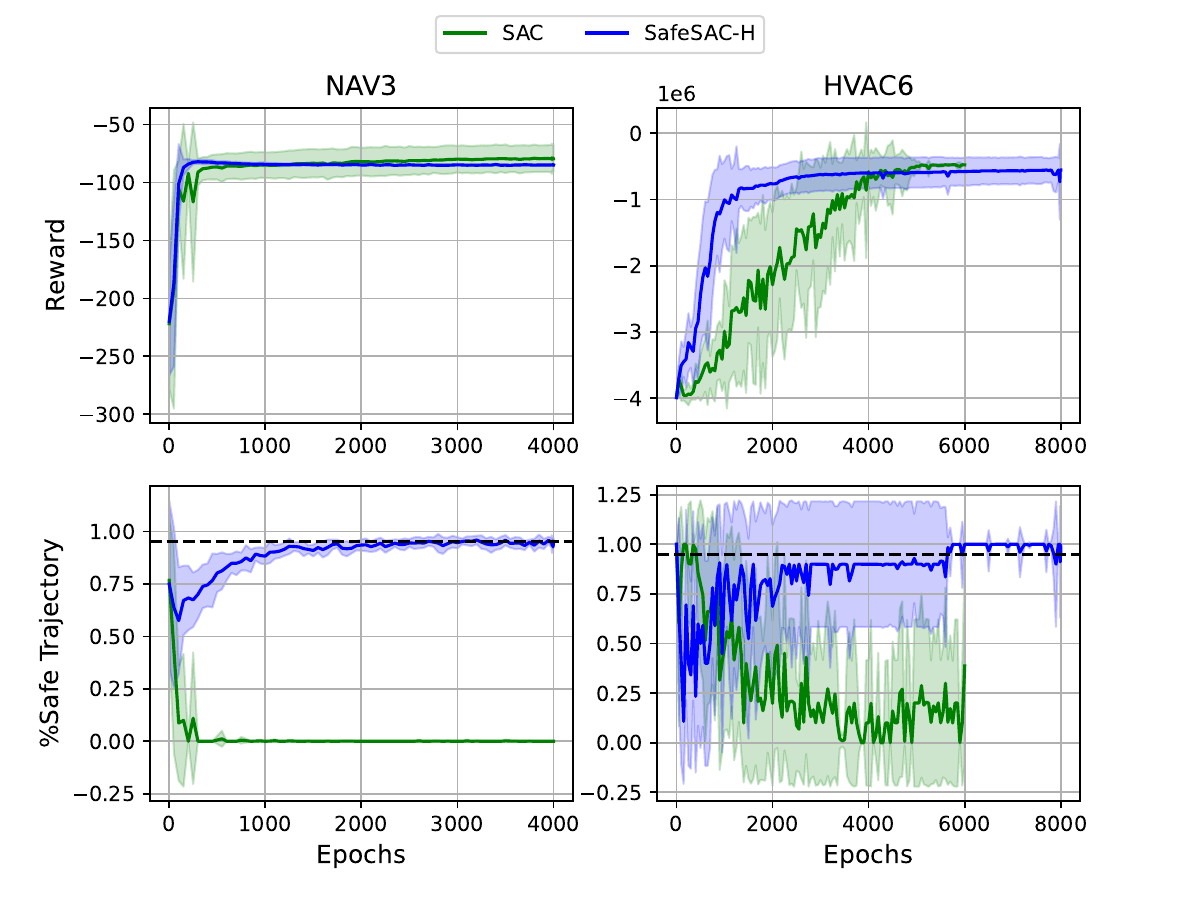}
    \vskip 0pt
    \caption{RDDL Results - Reward and Safety Performance}
    \label{fig:rddl_full}
    \vskip 0pt
\end{figure*}

\end{document}